\documentclass[letterpaper, 10 pt, conference]{ieeeconf}  

\IEEEoverridecommandlockouts                              

\let\labelindent\relax

\usepackage{url}
\usepackage{graphicx}
\usepackage{booktabs}
\usepackage{xcolor}
\usepackage{subfig}
\usepackage{cite}
\usepackage{comment}
\usepackage{balance}
\usepackage{enumitem}

\title{\LARGE \bf
Experimental Evaluation of ROS-Causal in Real-World \\Human-Robot Spatial Interaction Scenarios
}
\author{Luca Castri\textsuperscript{1}, Gloria Beraldo\textsuperscript{2,3}, Sariah Mghames\textsuperscript{1}, Marc Hanheide\textsuperscript{1}, Nicola Bellotto\textsuperscript{1,2}
\thanks{\hspace{-3mm}\textsuperscript{1}School of Computer Science, University of Lincoln, UK.\newline
\textsuperscript{2}Dept. of Information Engineering, University of Padua, Italy.\newline
\textsuperscript{3}National Research Council of Italy.\newline
This work has received funding from the European Union’s Horizon 2020 research
and innovation programme under grant agreement No 101017274 (DARKO).
GB is also supported by PNRR MUR project PE0000013-FAIR.}
}

\begin{document}

\maketitle
\thispagestyle{empty}
\pagestyle{empty}

\begin{abstract}
Deploying robots in human-shared environments requires a deep understanding of how nearby agents and objects interact. Employing causal inference to model cause-and-effect relationships facilitates the prediction of human behaviours and enables the anticipation of robot interventions. However, a significant challenge arises due to the absence of implementation of existing causal discovery methods within the ROS ecosystem, the standard de-facto framework in robotics, hindering effective utilisation on real robots. To bridge this gap, in our previous work we proposed ROS-Causal, a ROS-based framework designed for onboard data collection and causal discovery in human-robot spatial interactions. In this work, we present an experimental evaluation of ROS-Causal both in simulation and on a new dataset of human-robot spatial interactions in a lab scenario, to assess its performance and effectiveness. Our analysis demonstrates the efficacy of this approach, showcasing how causal models can be extracted directly onboard by robots during data collection. The online causal models generated from the simulation are consistent with those from lab experiments. These findings can help researchers to enhance the performance of robotic systems in shared environments, firstly by studying the causal relations between variables in simulation without real people, and then facilitating the actual robot deployment in real human environments.\\
ROS-Causal: {\footnotesize\url{https://lcastri.github.io/roscausal}}
\end{abstract}

\section{INTRODUCTION}
The increasing deployment of robots for industrial, agricultural, and healthcare applications can bring a significant advancement in these sectors. However, the integration of robots alongside humans requires novel approaches for effective human-robot interaction~(HRIs). When sharing a workspace with humans, robots must consider how their actions influence human behaviour.
Understanding the cause-effect relation between robot's actions and human reactions is crucial to enhance HRIs. Causal reasoning is a crucial step towards making the robot more efficient and safe. The study of these relations falls within causal inference~\cite{pearl2009causality} and starts with the actual {\em discovery} of such relations.

Knowing the causal structure of a process is desirable for many types of applications. Therefore, causal discovery and reasoning can be encountered in the literature across different fields, including of course robotics~\cite{brawer_causal_2021,cao_reasoning_2021,castri2022causal,castri2023enhancing,Katz2018,Angelov2019,Lee2022,cannizzaro2023towards,cannizzaro2023towardsdrones,cannizzaro2023car}. 
However, many causal discovery methods in these applications need a two-step process: real-world data collection, followed by {\em offline} causal analysis. This approach is not suitable for runtime operations and limits the capability of the robot to learn and reason ``on the field''.
%
Consider a scenario where a robot interacts with a person in a warehouse. Due to the aforementioned limitation, the robot is required to accumulate a substantial amount of data and then conduct offline causal analysis. Subsequently, the reconstructed causal model must be reintegrated into the robot for actual use.

\begin{figure}
  \includegraphics[trim={0.2cm 0.1cm 6.5cm 5.3cm},clip,width=\columnwidth]{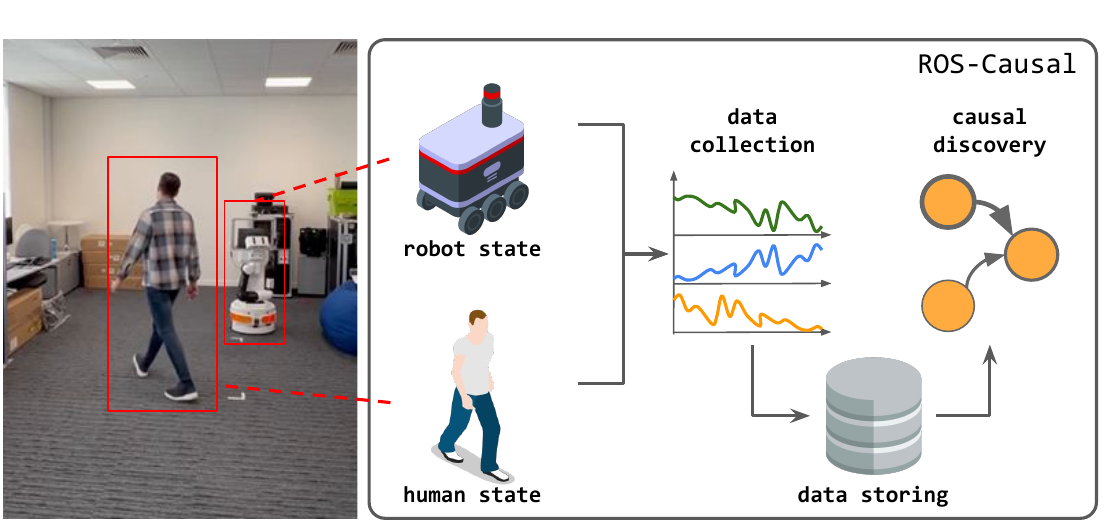}
  \caption{A high-level overview of the core components of ROS-Causal.}
  \label{fig:intro}
\end{figure}

One reason for such limitation is the lack of frameworks that facilitate the integration between the two research communities -- causal inference and robotics --  within ROS\footnote{\url{https://www.ros.org/}}(Robot Operating System), the de-facto standard software framework in robotics.
To this end, in our recent work we introduced {\em ROS-Causal}, a ROS-based framework designed for conducting onboard data collection and causal discovery during HRIs~\cite{castri2024ros}. ROS-Causal allows robots to analyse data batches while simultaneously collecting data for future causal analysis. A high-level representation of ROS-Causal is depicted in Fig.~\ref{fig:intro}. 

By integrating our framework within ROS, the acquired causal model can be directly exploited by the robot. In a preliminary study~\cite{castri2024ros}, we outlined the general properties of ROS-Causal and demonstrated its capabilities within {\em ROS-Causal\_HRISim}, a Gazebo-based robotic simulator tailored for the design of HRI scenarios and their causal analysis. Building upon our previous work, this paper presents a comprehensive experimental evaluation of ROS-Causal in a real-world lab environment, replicating and extending the human-robot spatial interactions~(HRSIs) originally simulated in~\cite{castri2024ros}. In summary, our contributions are as follows:
\begin{itemize}
    \item the first runtime creation of a HRSI causal model onboard the robot with its sensors data, via ROS-Causal;
    \item an experimental evaluation of the latter in HRSI scenarios, including 15 human participants;
    \item a new, publicly available dataset of human-goal and HRSI trajectories in the experimental scenario, used to evaluate and complement the ROS-Causal framework.
\end{itemize}

The paper is structured as follows: an overview of causal discovery methods and their applications in robotics is provided in Section~\ref{sec:related}; Section~\ref{sec:appr} explains the functionalities of ROS-Causal and the evaluation methodology; Section~\ref{sec:exp} presents the experimental settings and results; finally, Section~\ref{sec:conclusion} concludes the paper discussing achievements and future work.

\section{RELATED WORK}\label{sec:related}
\noindent\textbf{Causal discovery:} A variety of methods have been developed for causal discovery, aimed at inferring causal relationships from observational data. These methods are broadly classified into three categories~\cite{glymour_review_2019}: {\em (i) constraint-based methods}, like Peter \& Clark~(PC) and Fast Causal Inference~(FCI)~\cite{spirtes2000causation}; {\em (ii) score-based methods}, such as Greedy Equivalence Search~(GES) and NOTEARS~\cite{zheng2018dags}; and {\em (iii) noise-based methods}, like Linear Non-Gaussian Acyclic Models~(LiNGAM)~\cite{shimizu2006linear}.
However, many of these algorithms are designed solely for static data and may not be suitable for analysing time-series sensor data, which is common in robotics applications. In such cases, time-dependent causal discovery methods become essential.
Several algorithms for causal discovery from time-series data have been developed to address this need~\cite{assaad2022survey}.
Within the area of Granger causality, there is Temporal Causal Discovery Framework~(TCDF)~\cite{nauta2019causal}.
In \cite{pamfil2020dynotears}, the time-series version of NOTEARS, i.e. DYNOTEARS, is presented.
Among the noise-based methods, there are Time Series Models with Independent Noise~(TiMINo)~\cite{peters2013causal} and Vector Autoregressive LiNGAM~(VARLiNGAM)~\cite{hyvarinen2010estimation}. In the constraint-based category, variations of the FCI and PC algorithms, namely Time-series FCI~(tsFCI)~\cite{entner2010causal} and PC Momentary Conditional Independence~(PCMCI)~\cite{runge_causal_2018}, were tailored to handle time-series data. 
PCMCI, with wide applications in climate, healthcare, and robotics~\cite{runge_detecting_2019,saetia_constructing_2021,castri2022causal}, has recently seen extensions such as PCMCI\textsuperscript{+}~\cite{runge2020discovering} for discovering simultaneous dependencies and Filtered-PCMCI~(F-PCMCI)~\cite{castri2023enhancing}, which incorporates a transfer entropy-based feature-selection module to enhance causal discovery by focusing on relevant variables.

\noindent\textbf{Causal robotics:}
The synergy between causality and robotics offers mutual benefits. Causality leverages robots' physical capabilities for interventions, while robots utilise causal models to gain a deeper understanding of their environment. This synergy has led to increased attention towards causal inference in various robotics applications.
For instance, Structural Causal Models~(SCM) have been employed to comprehend how humanoid robots interact with tools~\cite{brawer_causal_2021}. PCMCI and F-PCMCI have been utilised to establish causal models for underwater robots reaching target positions~\cite{cao_reasoning_2021} and to predict human spatial interactions in social robotics~\cite{castri2022causal,castri2023enhancing}. Additionally, causality-based approaches have been explored in robot imitation learning, manipulation, drone applications, and planning~\cite{Katz2018,Angelov2019,Lee2022,cannizzaro2023towards,cannizzaro2023towardsdrones,cannizzaro2023car}.
However, current causal discovery methods typically entail offline analysis post-data collection and are not integrated into ROS, presenting challenges for their adoption and experimental reproducibility in robotics. Our objective was to develop a modular ROS-based causal analysis framework that enables simultaneous data collection and causal analysis processes. The proposed framework can be used in simulation and on real robots. Indeed, ROS-Causal is integrated with ROS-Causal\_HRISim~\cite{castri2024ros}, a Gazebo-based simulator tailored for HRI scenarios, facilitating the data collection and the interventions on both robots and humans, to verify the resulted causal model before involving people.

\begin{figure*}[t]
  \includegraphics[trim={0cm 0cm 4.5cm 9.3cm},clip,width=\textwidth]{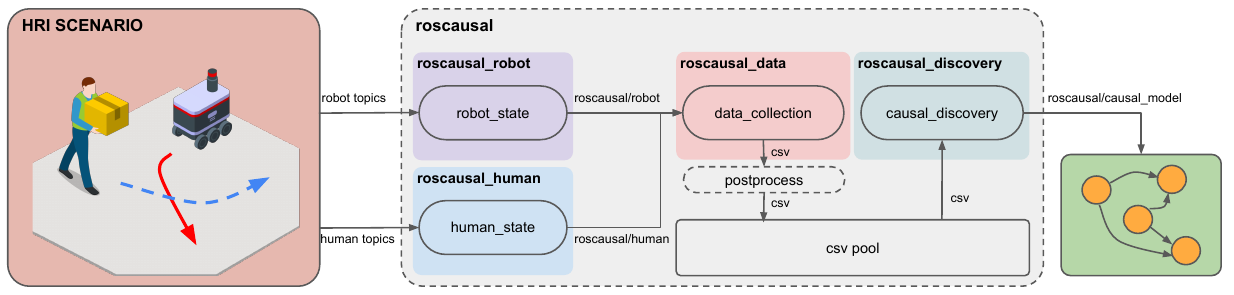}
  \caption{
  ROS-Causal pipeline~\cite{castri2024ros}: \emph{(i)} data extraction from HRI scenarios; \emph{(ii)} collection and post-processing of data to derive a high-level representation of the scenario; \emph{(iii)} causal discovery conducted on the extracted data, with the resulting causal model published on a dedicated rostopic.}
  \label{fig:roscausal}
\end{figure*}

\noindent\textbf{Human-robot spatial interaction:}
Several studies have proposed methods to model the 
relations between human motion behaviours and spatial interactions. 
In~\cite{Liu2022}, a high-level causal framework for motion forecasting is proposed. This framework incorporates human interactions and employs a dynamic process with various latent variables to account for unobserved and spurious features.
In social robotics, 
qualitative trajectory calculus (QTC) is used to account for motion relations between human-human and human-robot pairs~\cite{bellotto2013qualitative,Dondrup2014,hanheide2012analysis,mghames2023neuro,mghames2023qualitative}. However, causal relations between spatial variables have not been considered. This work builds upon our previous research~\cite{castri2022causal,castri2023enhancing}, and is inspired by the QTC relations, extended to incorporate additional factors like collisions, represented quantitatively rather than qualitatively.

\noindent\textbf{Human-robot spatial interaction dataset:} Various datasets capturing human-human and human-robot spatial interactions are available in the literature. 
The TH{\"O}R dataset~\cite{thorDataset2019} and its extension, TH{\"O}R-MAGNI~\cite{schreiter2022magni}, provide collections of human motion trajectories recorded from cameras installed on the ceiling of a controlled indoor environment.
The ATC Pedestrian Tracking dataset~\cite{brvsvcic2013person} offers trajectories of people in a large atrium of a shopping mall. Tracking was performed using several 3D range sensors installed on the ceiling.
The JackRabbot Dataset and Benchmark (JRDB)~\cite{martin2021jrdb} features data collected from the social mobile manipulator JackRabbot. This dataset includes RGB camera and 3D LiDAR data capturing human poses in both indoor and outdoor scenarios, from the ego-perspective of the robot, both stationary and navigating.
In contrast, our dataset focuses specifically on human-goal and human-robot spatial interaction in an indoor environment, captured from the perspective of a 3D Velodyne VLP-16 LiDAR mounted on the TIAGo robot. 
Furthermore, our dataset specifically emphasizes human spatial behaviors in reaching predefined target positions while avoiding interaction with the robot. This emphasis provides rich time-series data suitable for causal analysis.

\section{ROS-CAUSAL FOR HUMAN-ROBOT SPATIAL INTERACTION MODELLING}\label{sec:appr}
\subsection{ROS-Causal}\label{subsec:roscausal}
Our approach, named ROS-Causal~\cite{castri2024ros}, introduces a ROS-based causal analysis framework designed for extracting and collecting data from an HRI scenario, such as agents' trajectories, followed by conducting causal analysis on the collected data in a batched manner. ROS-Causal consists of three main components:
\begin{itemize}
    \item \textbf{Data Merging}: it comprises the ROS nodes \texttt{roscausal\_robot} and \texttt{roscausal\_human}. They collect data (e.g., position, velocity, target position, etc.) from several rostopics related to the robot and human, respectively, and then merge them into a single rostopic used within the framework;
    \item \textbf{Data Collection and Post-processing}: it includes the \texttt{roscausal\_data} node, which takes input from the merged topics and generates CSV data batches for the subsequent causal discovery node. Once the desired time-series length is reached, the node offers the option to post-process the data and then saves it into a dedicated folder (e.g., "csv\_pool" as in Fig.~\ref{fig:roscausal});
    \item \textbf{Causal Discovery}: this component encompasses the \texttt{roscausal\_discovery} node, responsible for performing causal discovery analysis on the CSV batch files generated by the previous component. The node publishes the obtained causal model in a rostopic. It supports two causal discovery methods: PCMCI~\cite{runge_causal_2018} and its extension, F-PCMCI~\cite{castri2023enhancing}.
\end{itemize}
It is important to note that the \texttt{roscausal\_data} and \texttt{roscausal\_discovery} nodes operate asynchronously, allowing the simultaneous execution of causal analysis on one dataset, while continuing the collection of another. 
%
In this work, the data are saved in CSV files that are generated by \texttt{roscausal\_data} and then provided in input to the \texttt{roscausal\_discovery}. This choice facilitates offline analysis and dataset creation. However, it is also possible to create a causal model ``on the fly'' based on the data collected through the rostopic.
%
The complete pipeline of ROS-Causal is depicted in Fig.~\ref{fig:roscausal}. A webpage with the Python implementation of ROS-Causal and both evaluation strategies (simulation as described in Sec.~\ref{subsec:roscausal_hrisim}, and real-world data as presented in Sec.~\ref{subsec:roscausal_real}), is publicly available\footnote{\url{https://lcastri.github.io/roscausal}}.

\subsection{Human-Robot Spatial Interaction Scenario}\label{subsec:hrsi}
By taking inspiration from the spatial interaction in a dynamic agile warehouse where different agents have to move to pick/place objects, we devised a HRSI scenario, that we have started to analyse in~\cite{castri2022causal} and herein, we reproduce with a real robot to assess ROS-Causal. A person and a robot deliver parcels at different target stations. The person has to reach a predefined target position, which dynamically changes when reached, and avoid the robot that crosses his/her path. The robot follows a predetermined path along its targets. When the person encounters the robot, he/she must avoid it by decreasing his/her velocity and/or adjusting his/her steering. 
In addition, as the person approaches the target position, he/she gradually reduces the velocity. To design the scenario, we took inspiration from the human's and robot's motions proposed in the TH{\"O}R dataset~\cite{thorDataset2019}.

The set of variables used to model this scenario aligns with those chosen in~\cite{castri2022causal} for causal analysis and includes the following: {\em (i)} $v$ - human velocity; {\em (ii)} $d_g$ - human's distance to his target position; {\em (iii)} $r$ - risk of collision with the robot. The expected causal links in this scenario are as follows:
\begin{itemize}
    \item $v \rightarrow d_g$: $d_g$ depends inversely on $v$;
    \item $d_g \rightarrow v \leftarrow r$: $v$ is a direct function of the $d_g$, but it is also affected by the collision $r$;
    \item $v \rightarrow r$: $r$ depends on the velocity, as explained in~\cite{castri2022causal}.
\end{itemize}
In this paper, the robot is perceived as an obstacle by the person. However, ROS-Causal can be further applied to various scenarios involving robots and humans, such as a robot following a person or interactive tasks between them. 

\section{EXPERIMENTS}\label{sec:exp}

\begin{figure*}[t]
\centering
\includegraphics[trim={0cm 0cm 0cm 6cm}, clip, width=\textwidth]{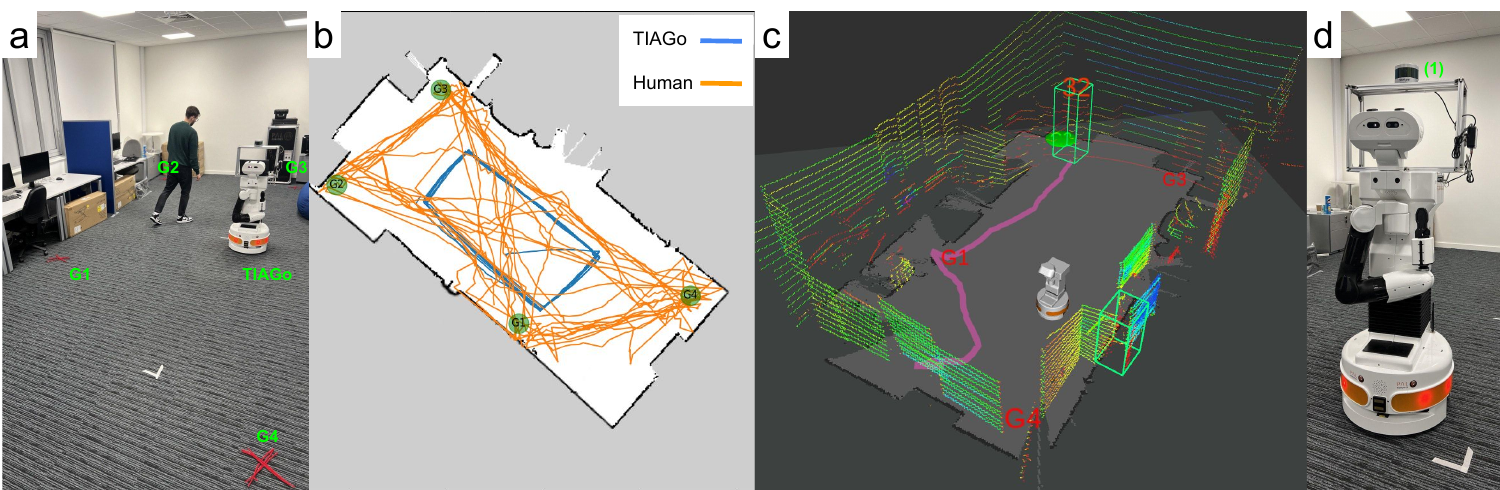}
\caption{(a) HRSI experiment in a lab scenario with a TIAGo robot, a person and his/her four goal positions;(b) 2D map of an experiment with a person and TIAGo, with trajectories in orange and blue respectively, and four goal positions (green dot); (c) RViz visualisation of the scenario; (d) TIAGo robot with (1) a Velodyne VLP-16 3D LiDAR used for dataset collection.}
\label{fig:exp_setup}
\end{figure*}

\begin{figure}[t]
\centering
\includegraphics[width=\columnwidth]{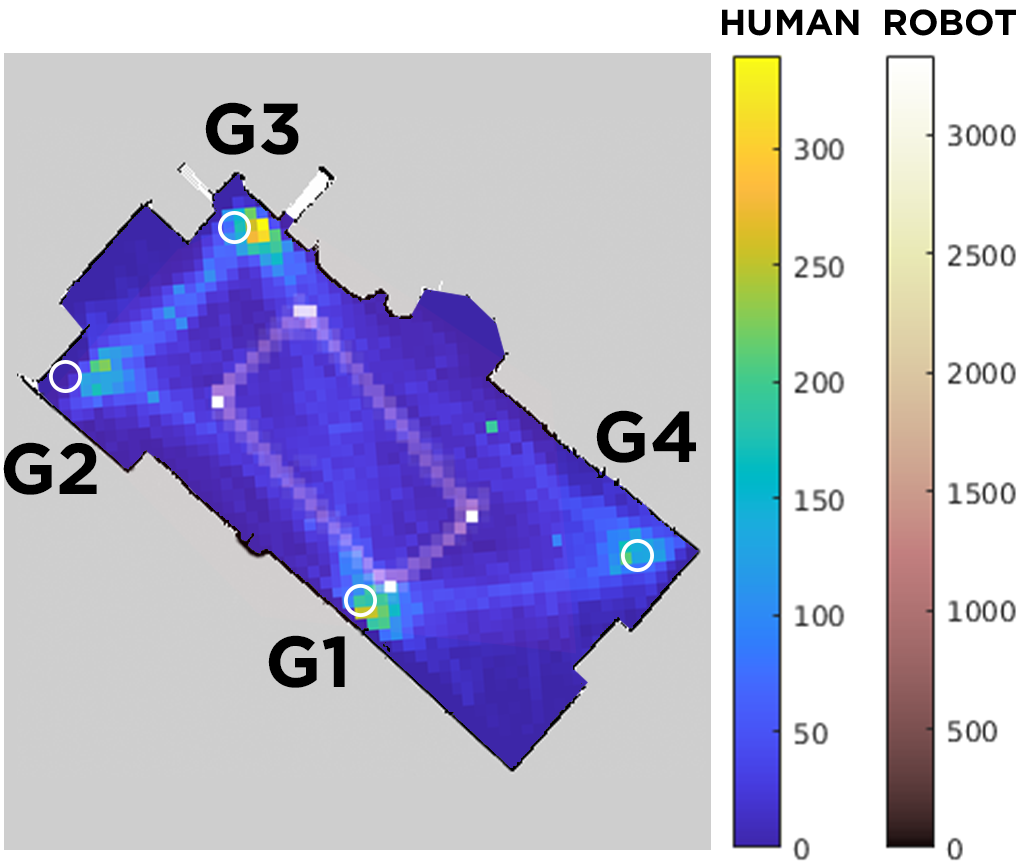}
\caption{Heat map of the all participants (left colormap) and robot (right colormap) trajectories during the experiments. The color palette varies based on the hit rate per cell.}
\label{fig:exp_heatmap}
\end{figure}

Our evaluation strategy consists of two steps. First, we validate the correctness and effectiveness of ROS-Causal in a simulated HRSI environment. This step is crucial for assessing ROS-Causal's capability to reconstruct the correct causal model from data before deploying it on the real robot. Second, we evaluate ROS-Causal in a real HRSI scenario, where data collection and causal discovery are performed directly on the real robot. The experiments have been designed to investigate the following research questions:
\begin{enumerate}[label={R$_{\arabic*}$)}]
    \item Is it feasible to generate causal models onboard the robot via ROS-Causal?
    \item If yes,
     how much data (i.e., time-series length and sampling frequency) are needed to generate accurate causal models? 
    \item If yes, how much execution time the generation takes?
\end{enumerate}

\subsection{ROS-Causal Simulation Evaluation}\label{subsec:roscausal_hrisim}
In our previous work~\cite{castri2024ros}, we assess ROS-Causal's effectiveness in reconstructing causal models from HRSI scenarios using ROS-Causal\_HRISim, a dedicated Gazebo-based simulator. This simulator replicates HRI scenarios involving a TIAGo\footnote{\url{https://pal-robotics.com/robots/tiago/}} robot and multiple pedestrians simulated using the {\em pedsim\_ros}\footnote{\url{https://github.com/srl-freiburg/pedsim_ros}} ROS library. The latter simulates individual and group social activities (e.g., walking) using a social force model. To better emulate human behaviours, we included user teleoperation (via keyboard) of a simulated person, unaffected by social forces.
%

\begin{figure}[!h]\centering
\includegraphics[trim={6cm 5cm 6cm 4cm}, clip, width=0.7\columnwidth]{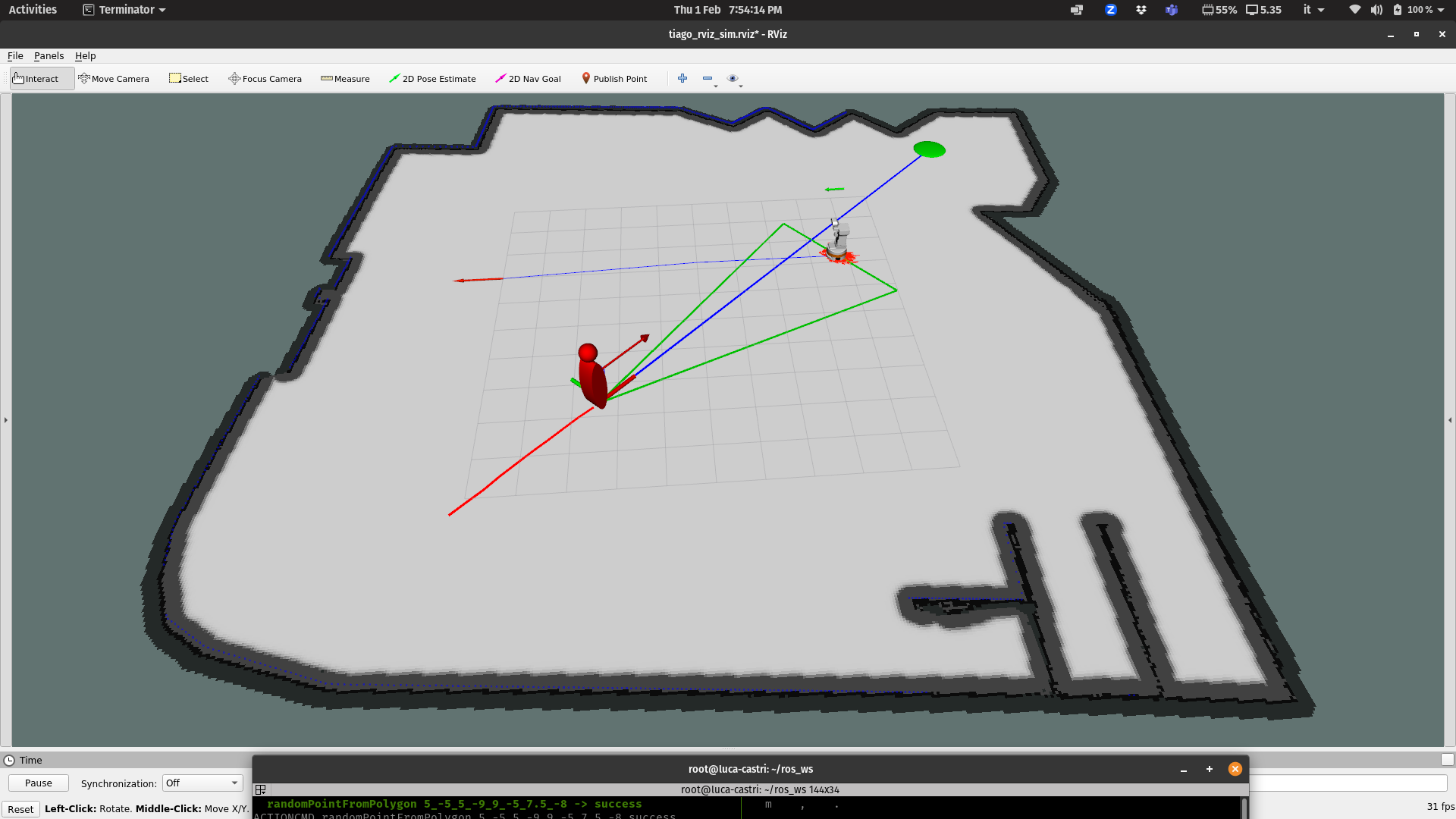}
\caption{HRI scenario involving a TIAGo robot and a teleoperated person, created by ROS-Causal\_HRISim.}
\label{fig:exp_causal_hri}
\end{figure}

Our plan was to create the HRSI scenario presented in Section~\ref{subsec:hrsi}, collect the trajectories of the two agents (i.e., robot and person), process the collected data to obtain the desired set of variables previously discussed $(v, d_g, r)$ and finally execute the causal discovery on it. Fig.~\ref{fig:exp_causal_hri} shows the HRSI scenario created by ROS-Causal\_HRISim. 

It involves a TIAGo robot and a simulated person teleoperated by one of the participants via keyboard, represented by the red manikin. The green dot represents the person's target position, while the blue line visualises the distance between the person and his/her goal position. Finally, the green cone is the visualisation of the collision risk. It is built from the person position to the enlarged encumbrance of the TIAGo, which is perceived by the human as a moving obstacle.

\begin{figure}[!h]\centering
\subfloat[]{
\label{fig:exp_simulation}\includegraphics[trim={6.6cm 2.2cm 5.3cm 1cm}, clip, width=0.48\columnwidth]{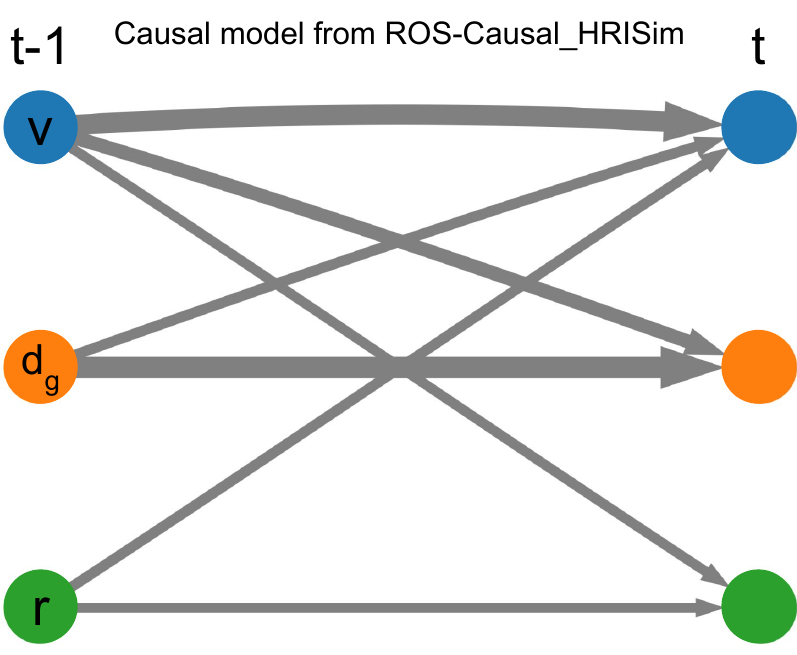}}
\subfloat[]{
    \label{fig:exp_realworld}
    \includegraphics[trim={6.6cm 2.2cm 5.3cm 1cm}, clip, width=0.48\columnwidth]{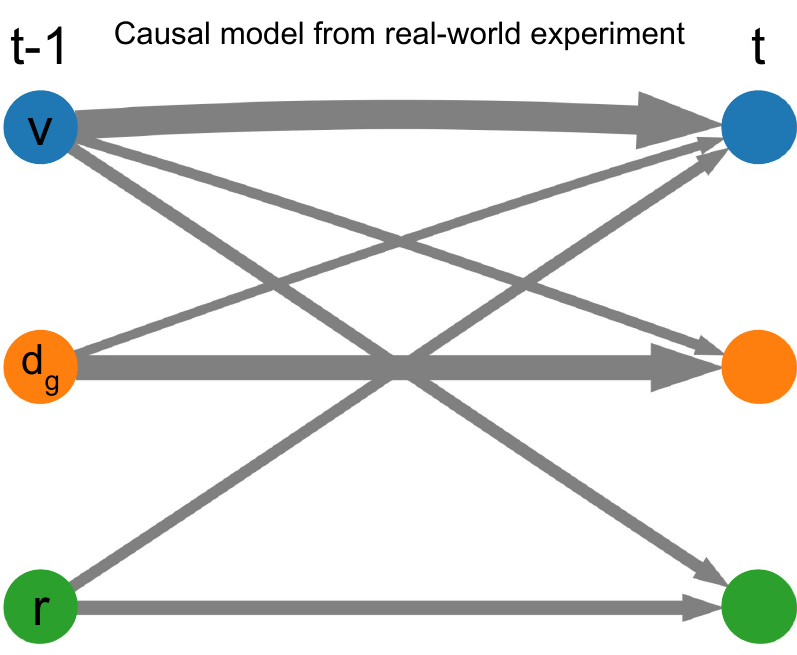}}\\
\subfloat[]{
    \label{fig:exp_freq}
    \includegraphics[trim={0.5cm 0cm 1cm 1.2cm}, clip, width=0.84\columnwidth]{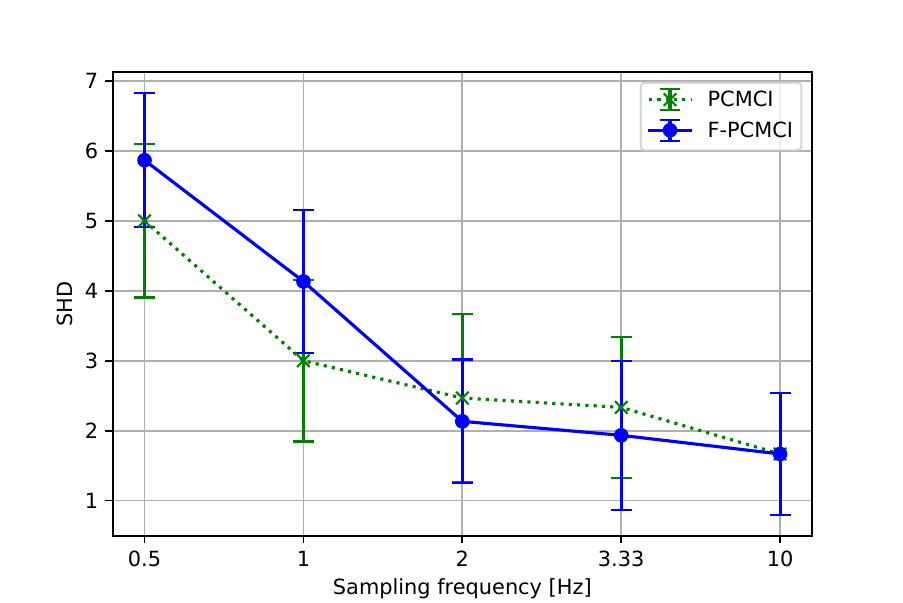}}\\
\subfloat[]{
\label{fig:exp_shd}\includegraphics[trim={0.6cm 0cm 1cm 1.2cm}, clip, width=0.84\columnwidth]{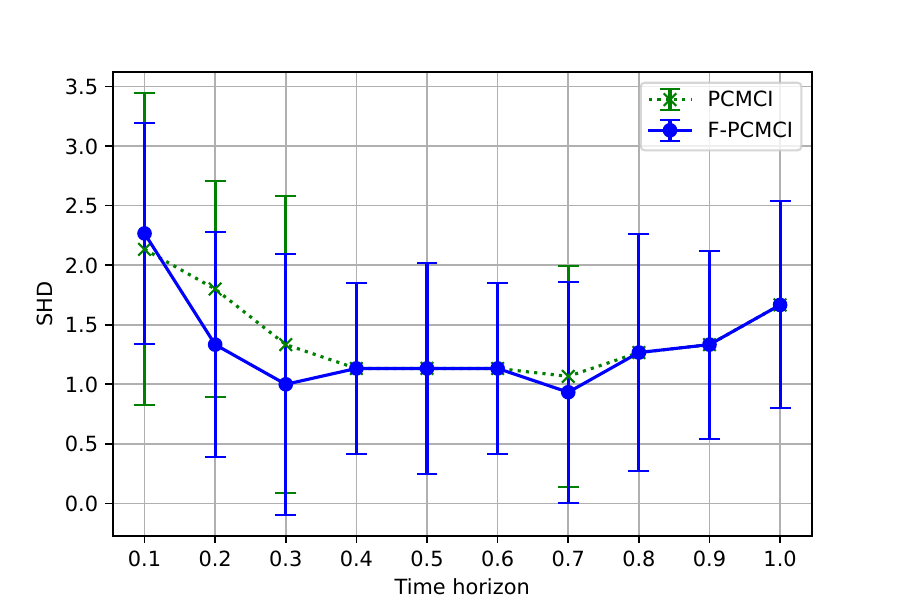}}\\
\subfloat[]{
    \label{fig:exp_time}
    \includegraphics[trim={0.5cm 0cm 1cm 1.2cm}, clip, width=0.84\columnwidth]{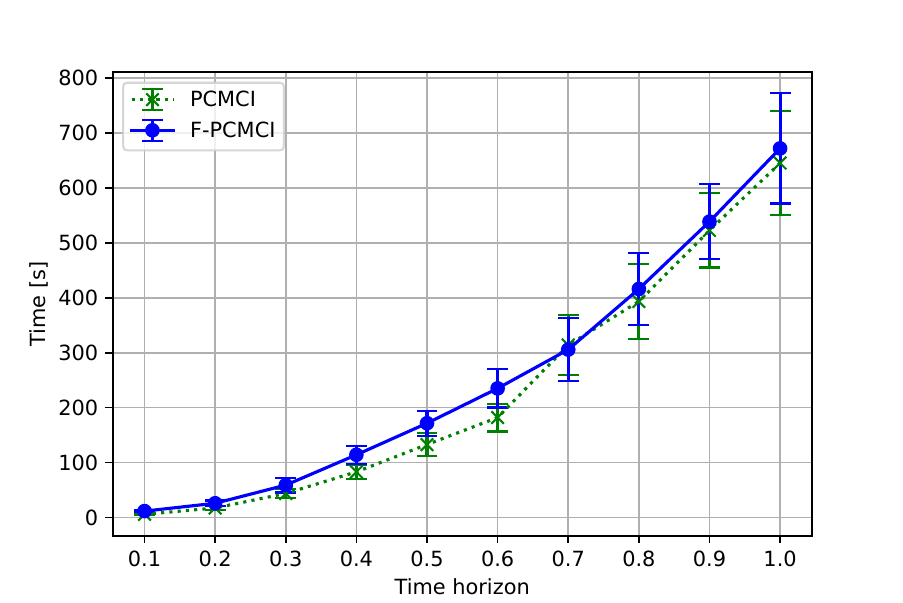}}\\
\caption{(a) and (b) Causal model reconstructed by ROS-Causal from simulation and lab experiments, respectively. Execution time (c) and SHD (d) analyses with various time horizons. (e) SHD analysis based on the sampling frequency.}
\end{figure}

Regarding the ROS-Causal parameters and settings used for the data collection and causal analysis, we configured a desired time-series length corresponding to a timeframe of $150$s and recorded the trajectories of the two agents, their linear velocity, and orientation, with a sampling frequency of $10$Hz. Subsequently, we compute
the distance between the human and the goal, as well as the risk of collision. For the causal discovery block, we employed the F-PCMCI causal discovery method with a significance level of $\alpha = 0.05$, a conditional independence test based on Gaussian Process regression and Distance Correlation (GPDC). We also used a 1-step lag time, meaning variables at time $t$ could only be affected by those at time $t-1$. The resulting causal model is depicted in Fig.~\ref{fig:exp_simulation}.
The graph faithfully represents the expected model discussed earlier and is consistent with the results in~\cite{castri2022causal}.
    
\subsection{Dataset and Experimental Setup}

After confirming the correct functionalities of the ROS-Causal framework, we proceeded with the lab evaluation analysis, replicating the scenario staged through ROS-Causal\_HRISim in the lab environment, as shown in Fig.~\ref{fig:exp_setup}a.
The experiment and data collection occurred in a laboratory room of $5\times8.2m$ in Fig.~\ref{fig:exp_setup}b. Fifteen participants (6 females), aged between 25 and 55, took part in the experiment. Seven of them were researchers who regularly work with robots. 
All participants voluntarily agreed to take part in the experiments and signed a written informed consent.
Only point cloud readings from the Velodyne VLP-16 3D LiDAR were recorded, without any sensitive data (such as faces). 

Participants' task was to navigate between four designated goal positions while avoiding collisions with the robot when crossing paths. Specifically, they were instructed to begin from a goal position randomly chosen by themselves, select and walk towards the next one, also randomly chosen, and repeat this process until the robot stopped (i.e., after 5 minutes from the start). They were asked to pass through all the goal positions at least 7 times, avoiding the robot when they encountered it. No specific instructions were provided on how to reach the goals or avoid the robot.

A predefined rectangular path (i.e., in blue in Fig.~\ref{fig:exp_setup}b) was set for the TIAGo robot to navigate along the room and generate frequent interactions with the participants.
%
As mentioned earlier, in this experimental setting, the robot was considered by the participant as an obstacle to avoid while walking towards their target positions. 
Fig.~\ref{fig:exp_setup}a shows an example of the experiment, while Fig.~\ref{fig:exp_setup}b shows the trajectories of the two agents (i.e., the ones related to the robot in blue and the human in orange).

%

As already mentioned, to track the motion of the agents, we used a Velodyne VLP-16 3D LiDAR and the Bayes People Tracker\footnote{\url{https://github.com/LCAS/bayestracking}}~\cite{yan2017online} on the related point cloud. This LiDAR features 16 scan channels, providing a $360^{\circ}$ horizontal and $30^{\circ}$ vertical field-of-view. Fig.~\ref{fig:exp_setup}c illustrates, through RViz, the human tracked by the robot.
The robot equipped with the Velodyne as shown in Fig.~\ref{fig:exp_setup}d was configured to navigate with its torso positioned at the minimum height. This arrangement ensured that the Velodyne was positioned at a height of $1.2m$ from the floor, maximising the chance to detect individuals even when they were close to the robot. The Velodyne LiDAR was set to rotate at 10 Hz. 

The dataset collection and all experiments reported in this paper were conducted using Ubuntu 20.04 LTS (64-bit) and ROS Noetic, running on an Intel i7-10875H processor with 16 GB memory. 
The dataset consists of both rosbag and CSV files.
The rosbag includes topics related to the map of the laboratory room, robot state (position, orientation, and velocities), the Velodyne point cloud data, and goal positions. 
Additionally, the dataset includes a CSV file per participant containing all processed time-series data. 
Fig.~\ref{fig:exp_heatmap} shows the heat map (cell resolution equal to 0.15m) related to the distribution of all the participants (blue colormap) and the robot (pink colormap) trajectories collected during the experiments. The figure highlights the variance of the walking gaits over the involved participants. As expected, the areas near the target goals are the most shared among the different trajectories. 

\subsection{ROS-Causal Evaluation in Lab Scenario}\label{subsec:roscausal_real}
Data collection, post-processing, and causal discovery were all executed by our ROS-Causal framework with the same parameters used for the simulation, as explained in Section~\ref{subsec:roscausal_hrisim}.
Fig.~\ref{fig:exp_realworld} shows the causal model relative to one of the participants. 
The graph accurately represents the expected model discussed in Section~\ref{subsec:hrsi} and is consistent with the results presented in~\cite{castri2022causal}, as well as with the model obtained from the simulation experiment in Fig.~\ref{fig:exp_simulation}. This demonstrates the reliability of ROS-Causal\_HRISim to mimic HRI scenarios
and the ROS-Causal's ability to retrieve the expected causal model in both simulation and lab experiments,  validating the onboard causal discovery via ROS-Causal~(R$_1$).

Fig.~\ref{fig:exp_freq},~\ref{fig:exp_shd} and~\ref{fig:exp_time} present the data requirement analyses, using the Structural Hamming Distance (SHD) and the execution time as metrics. These analyses explore various data configurations, including time horizon (time-series length) and sampling frequency. The SHD is a simple metric that quantifies the difference in terms of number of edges to compare two causal graphs. In all the three analyses, we execute the causal analysis using both PCMCI and F-PCMCI causal discovery methods included in ROS-Causal.

Fig.~\ref{fig:exp_freq} analyses the sampling frequency. Using the full-length time-series data, we conduct the causal discovery analysis for various sampling frequencies, ranging from $0.5$Hz to the original $10$Hz sampling frequency. We measure the SHD of the retrieved causal model compared to our baseline.
Fig.~\ref{fig:exp_shd},~\ref{fig:exp_time} depict the SHD and the execution time metrics for different time horizons. Specifically, we consider different percentages of the time-series length, ranging from $10\%$ up to $100\%$ of the full length, corresponding to an average of $5$ minutes per participant. For each time-horizon, we measure the SHD of the reconstructed causal model with respect to the causal graph obtained in simulation (Fig.~\ref{fig:exp_simulation}), serving as a baseline and the execution time needed for the causal discovery analysis.
The three analyses were conducted for all the 15 experiment participants. The F-PCMCI line (blue) and PCMCI dashed line (green) represent the mean SHD across the 15 participants, while the error bars correspond to the standard deviation.

These analyses helped us identify
the sampling frequency and time-series length required to generate accurate causal models for this specific HRSI scenario, while also evaluating the execution time for the causal analysis. From Fig.~\ref{fig:exp_freq}, it is evident that the sampling frequency plays a crucial role in obtaining the correct causal model, with the original sampling frequency being the most appropriate for this scenario. Additionally, Fig.~\ref{fig:exp_shd} reveals the appropriate time-series length for learning the desired causal model. The window from $30\%$ to $70\%$ of the full length, corresponding to the timeframe of $90$s-$210$s, enables ROS-Causal to retrieve the correct causal model. Observing the scenario for less than $90$s does not provide enough information for ROS-Causal to learn the correct causal model. Conversely, having time-series data longer than $210$s can lead to overfitting of the parametric kernel estimator used by the causal discovery methods to perform the conditional independence tests.
In conclusions, using a $40\%$ ($120$s) length of the time-series recorded at $10$Hz appears to provide the best trade-off between accuracy of the causal model and time required to reconstruct it ($\sim100$s), which answers our research questions~(R$_2$) and~(R$_3$). 
In the F-PCMCI framework, we employed a non-parametric Kraskov estimator for the Transfer Entropy (TE)-based feature selection module instead of the Gaussian estimator. While the Gaussian estimator is faster, it assumes that the analysed time-series follows a Gaussian distribution, which was not valid in our scenario. Consequently, the execution time advantage of F-PCMCI over PCMCI, as observed in~\cite{castri2023enhancing}, is not visible in this case, and the two methods exhibit similar execution times.

\section{CONCLUSION}\label{sec:conclusion}
In this work, we evaluated the effectiveness of the ROS-Causal framework in modelling human-robot spatial interactions, both in simulated and lab environments. We first designed the same HRSI scenario in ROS-Causal\_HRISim and lab settings. We then performed causal discovery with ROS-Causal on both of them, obtaining consistent causal models. Our outcome shows the feasibility of onboard causal discovery with a real robot, and validates the simulator's ability to represent realistic HRI scenarios. Additionally, we demonstrated how to analyse execution time and data requirements (time-series length and sampling frequency) of a specific scenario for generating accurate causal models.

Our experimental setting in a lab environment was designed to test and validate ROS-Causal in real HRSI scenarios. Future work will investigate more complex interactions in logistics and similar working environments, where multiple people share the space with the robot. 
Another interesting direction would be to conduct a cause-effect estimation between variables to compare not only the structure of the retrieved causal model but also its actual parameters, such as causal link strengths.
Moreover, since the execution time is critical for many robotics applications, we will further improve our work on efficient causal discovery~\cite{castri2023enhancing} to speed up the process for large and complex models.
Finally, we plan to extend ROS-Causal's capabilities beyond causal discovery, especially to leverage causal models for tasks such as robot planning and real-time interaction prediction. This would further boost ROS-Causal's potential to be widely used for robotics research and industrial applications. 

\section*{ACKNOWLEDGEMENT}
The authors would like to thank Chris Waltham for his assistance in designing the support frame used to mount the Velodyne VLP-16 3D LiDAR on the TIAGo robot.

\bibliographystyle{IEEEtran}
\balance
\bibliography{IEEEabrv,references}

@article{glymour_review_2019,
  title={Review of causal discovery methods based on graphical models},
  author={Glymour, Clark and Zhang, Kun and Spirtes, Peter},
  journal={Frontiers in genetics},
  volume={10},
  pages={524},
  year={2019},
  publisher={Frontiers Media SA}
}

@article{zheng2018dags,
  title={Dags with no tears: Continuous optimization for structure learning},
  author={Zheng, Xun and Aragam, Bryon and Ravikumar, Pradeep K and Xing, Eric P},
  journal={Advances in neural information processing systems},
  volume={31},
  year={2018}
}

@article{shimizu2006linear,
  title={A linear non-Gaussian acyclic model for causal discovery.},
  author={Shimizu, Shohei and Hoyer, Patrik O and Hyv{\"a}rinen, Aapo and Kerminen, Antti and Jordan, Michael},
  journal={Journal of Machine Learning Research},
  volume={7},
  number={10},
  year={2006}
}

@article{saetia_constructing_2021,
  title={Constructing brain connectivity model using causal network reconstruction approach},
  author={Saetia, Supat and Yoshimura, Natsue and Koike, Yasuharu},
  journal={Frontiers in Neuroinformatics},
  volume={15},
  pages={619557},
  year={2021},
  publisher={Frontiers Media SA}
}

@inproceedings{castri2022causal,
  title={Causal discovery of dynamic models for predicting human spatial interactions},
  author={Castri, Luca and Mghames, Sariah and Hanheide, Marc and Bellotto, Nicola},
  booktitle={Int. Conf. on Social Robotics},
  pages={154--164},
  year={2022},
  organization={Springer}
}

@article{assaad2022survey,
  title={Survey and evaluation of causal discovery methods for time series},
  author={Assaad, Charles K and Devijver, Emilie and Gaussier, Eric},
  journal={Journal of Artificial Intelligence Research},
  volume={73},
  pages={767--819},
  year={2022}
}

@article{nauta2019causal,
  title={Causal discovery with attention-based convolutional neural networks},
  author={Nauta, Meike and Bucur, Doina and Seifert, Christin},
  journal={Machine Learning and Knowledge Extraction},
  volume={1},
  number={1},
  pages={19},
  year={2019},
  publisher={MDPI}
}

@article{hyvarinen2010estimation,
  title={Estimation of a structural vector autoregression model using non-gaussianity.},
  author={Hyv{\"a}rinen, Aapo and Zhang, Kun and Shimizu, Shohei and Hoyer, Patrik O},
  journal={Journal of Machine Learning Research},
  volume={11},
  number={5},
  year={2010}
}

@inproceedings{pamfil2020dynotears,
  title={Dynotears: Structure learning from time-series data},
  author={Pamfil, Roxana and Sriwattanaworachai, Nisara and Desai, Shaan and Pilgerstorfer, Philip and Georgatzis, Konstantinos and Beaumont, Paul and Aragam, Bryon},
  booktitle={Int. Conf. on Artificial Intelligence and Statistics},
  pages={1595--1605},
  year={2020},
  organization={PMLR}
}

@inproceedings{castri2023enhancing,
  title={Enhancing Causal Discovery from Robot Sensor Data in Dynamic Scenarios},
  author={Castri, Luca and Mghames, Sariah and Hanheide, Marc and Bellotto, Nicola},
  booktitle={2nd Conf. on Causal Learning and Reasoning},
  year={2023}
}

@article{entner2010causal,
  title={On causal discovery from time series data using FCI},
  author={Entner, Doris and Hoyer, Patrik O},
  journal={Probabilistic graphical models},
  pages={121--128},
  year={2010}
}

@article{runge_causal_2018,
  title={Causal network reconstruction from time series: From theoretical assumptions to practical estimation},
  author={Runge, Jakob},
  journal={Chaos: An Interdisciplinary Journal of Nonlinear Science},
  volume={28},
  number={7},
  year={2018},
  publisher={AIP Publishing}
}

@article{runge_detecting_2019,
  title={Detecting and quantifying causal associations in large nonlinear time series datasets},
  author={Runge, Jakob and Nowack, Peer and Kretschmer, Marlene and Flaxman, Seth and Sejdinovic, Dino},
  journal={Science advances},
  volume={5},
  number={11},
  pages={eaau4996},
  year={2019},
  publisher={American Association for the Advancement of Science}
}

@inproceedings{runge2020discovering,
  title={Discovering contemporaneous and lagged causal relations in autocorrelated nonlinear time series datasets},
  author={Runge, Jakob},
  booktitle={Conf. on Uncertainty in Artificial Intelligence (UAI)},
  pages={1388--1397},
  year={2020},
  organization={PMLR}
}

@article{peters2013causal,
  title={Causal inference on time series using restricted structural equation models},
  author={Peters, Jonas and Janzing, Dominik and Sch{\"o}lkopf, Bernhard},
  journal={Advances in neural information processing systems},
  volume={26},
  year={2013}
}

@inproceedings{cannizzaro2023towards,
  title={Towards a Causal Probabilistic Framework for Prediction, Action-Selection \& Explanations for Robot Block-Stacking Tasks},
  author={Cannizzaro, Ricardo and Routley, Jonathan and Kunze, Lars},
  booktitle={IEEE/RSJ Int. Conf. on Intell. Robots \& Systems (IROS) Workshop on Causality for Robotics},
  year={2023},
  month={},
  organization={IEEE}
}

@inproceedings{cannizzaro2023towardsdrones,
  title={Towards Probabilistic Causal Discovery, Inference \& Explanations for Autonomous Drones in Mine Surveying Tasks},
  author={Cannizzaro, Ricardo and Howard, Rhys and Lewinska, Paulina and Kunze, Lars},
  booktitle={IEEE/RSJ Int. Conf. on Intell. Robots \& Systems (IROS) Workshop on Causality for Robotics},
  year={2023},
  month={},
  organization={IEEE}
}

@inproceedings{cannizzaro2023car,
  title={CAR-DESPOT: Causally-Informed Online POMDP Planning for Robots in Confounded Environments},
  author={Cannizzaro, Ricardo and Kunze, Lars},
  booktitle={IEEE/RSJ Int. Conf. on Intell. Robots \& Systems (IROS)},
  year={2023},
  month={October},
  organization={IEEE}
}

@inproceedings{brawer_causal_2021,
  title={A causal approach to tool affordance learning},
  author={Brawer, Jake and Qin, Meiying and Scassellati, Brian},
  booktitle={IEEE/RSJ Int. Conf. on Intell. Robots \& Systems (IROS)},
  pages={8394--8399},
  year = {2020}
}

@INPROCEEDINGS{cao_reasoning_2021,
  author={Cao, Yu and Li, Boyang and Li, Qian and Stokes, Adam and Ingram, David and Kiprakis, Aristides},
  booktitle={2021 IEEE Int. Conf. on Robotics and Automation (ICRA)}, 
  title={Reasoning Operational Decisions for Robots via Time Series Causal Inference}, 
  year={2021},
  volume={},
  number={},
  pages={6124-6131},
  doi={10.1109/ICRA48506.2021.9561659}}

@INPROCEEDINGS{Lee2022,  
    author={Lee, Tabitha E. and Zhao, Jialiang Alan and Sawhney, Amrita S. and Girdhar, Siddharth and Kroemer, Oliver},  
    booktitle={2021 IEEE Int. Conf. on Robotics and Automation (ICRA)},   
    title={Causal Reasoning in Simulation for Structure and Transfer Learning of Robot Manipulation Policies}, year={2021},  
    volume={},  
    number={},  
    pages={4776-4782},  doi={10.1109/ICRA48506.2021.9561439}
}

@article{Katz2018,
    author = {Katz, Garrett and Huang, Di Wei and Hauge, Theresa and Gentili, Rodolphe and Reggia, James},
    journal = {IEEE Trans. on Cognitive and Developmental Systems},
    title = {{A novel parsimonious cause-effect reasoning algorithm for robot imitation and plan recognition}},
    year = {2018},
}

@inproceedings{Angelov2019,
    author = {Angelov, Daniel and Hristov, Yordan and Ramamoorthy, Subramanian},
    booktitle = {Proc. of the Int. Joint Conf. on Autonomous Agents and Multiagent Systems, AAMAS},
    title = {{Using causal analysis to learn specifications from task demonstrations}},
    year = {2019},
}

@book{spirtes2000causation,
  title={Causation, prediction, and search},
  author={Spirtes, Peter and Glymour, Clark N and Scheines, Richard},
  year={2000},
  publisher={MIT press}
}

@book{pearl2009causality,
  title={Causality},
  author={Pearl, Judea},
  year={2009},
  publisher={Cambridge university press}
}

@inproceedings{castri2024ros,
  title={ROS-Causal: A ROS-based Causal Analysis Framework for Human-Robot Interaction Applications},
  author={Castri, Luca and Beraldo, Gloria and Mghames, Sariah and Hanheide, Marc and Bellotto, Nicola},
  booktitle={Workshop on Causal Learning for Human-Robot Interaction (Causal-HRI), ACM/IEEE International Conference on Human-Robot Interaction (HRI)},
  year={2024}
}

@inproceedings{mghames2023neuro,
  title={A neuro-symbolic approach for enhanced human motion prediction},
  author={Mghames, Sariah and Castri, Luca and Hanheide, Marc and Bellotto, Nicola},
  booktitle={2023 Int. Joint Conf. on Neural Networks (IJCNN)},
  pages={1--8},
  year={2023},
  organization={IEEE}
}

@inproceedings{mghames2023qualitative,
  title={Qualitative Prediction of Multi-Agent Spatial Interactions},
  author={Mghames, Sariah and Castri, Luca and Hanheide, Marc and Bellotto, Nicola},
  booktitle={2023 32nd IEEE Int. Conf. on Robot and Human Interactive Communication (RO-MAN)},
  pages={1170--1175},
  year={2023},
  organization={IEEE}
}

@InProceedings{Liu2022,
author = {Y. Liu and R. Cadei and J. Schweizer and S. Bahmani and A. Alahi},
title = {Towards Robust and Adaptive Motion Forecasting: A Causal Representation Perspective},
booktitle = {Proc. of the IEEE/CVF Conf. on Computer Vision and Pattern Recognition (CVPR)},
year = {2022},
pages = {17081-17092},
}

@inproceedings{bellotto2013qualitative,
  title={Qualitative design and implementation of human-robot spatial interactions},
  author={Bellotto, Nicola and Hanheide, Marc and Van de Weghe, Nico},
  booktitle={Int. Conf. on Social Robotics},
  pages={331--340},
  year={2013},
  organization={Springer}
}

@inproceedings{hanheide2012analysis,
  title={Analysis of human-robot spatial behaviour applying a qualitative trajectory calculus},
  author={Hanheide, Marc and Peters, Annika and Bellotto, Nicola},
  booktitle={2012 IEEE RO-MAN: The 21st IEEE Int. Symposium on Robot and Human Interactive Communication},
  pages={689--694},
  year={},
  organization={}
}

@inproceedings{Dondrup2014,
  title={A probabilistic model of human-robot spatial interaction using a qualitative trajectory calculus},
  author={Dondrup, Christian and Bellotto, Nicola and Hanheide, Marc},
  booktitle={2014 AAAI Spring Symposium Series},
}

@article{thorDataset2019,
    title={TH{\"O}R: Human-Robot Navigation Data Collection and Accurate Motion Trajectories Dataset},
    author={Rudenko, Andrey and Kucner, Tomasz P and Swaminathan, Chittaranjan S and Chadalavada, Ravi T and Arras, Kai O and Lilienthal, Achim J},
    journal={IEEE Robotics \& Automation Letters},
    year={2020},
    publisher={IEEE},
    pages={676--682},
}

@inproceedings{schreiter2022magni,
  title={The Magni Human Motion Dataset: Accurate, Complex, Multi-Modal, Natural, Semantically-Rich and Contextualized},
  author={Schreiter, Tim and Almeida, Tiago Rodrigues de and Zhu, Yufei and Guti{\'e}rrez Maestro, Eduardo and Morillo-Mendez, Lucas and Rudenko, Andrey and Kucner, Tomasz P and Martinez Mozos, Oscar and Magnusson, Martin and Palmieri, Luigi and others},
  booktitle={31st IEEE Int. Conf. on Robot \& Human Interactive Communication},
  year={2022}
}

@article{brvsvcic2013person,
    title={Person tracking in large public spaces using 3-D range sensors},
    author={Br{\v{s}}{\v{c}}i{\'c}, Dra{\v{z}}en and Kanda, Takayuki and Ikeda, Tetsushi and Miyashita, Takahiro},
    journal={IEEE Trans. on Human-Machine Systems},
    year={2013},
    publisher={IEEE},
    pages={522--534},
}

@article{martin2021jrdb,
    title={Jrdb: A dataset and benchmark of egocentric robot visual perception of humans in built environments},
    author={Martin-Martin, Roberto and Patel, Mihir and Rezatofighi, Hamid and Shenoi, Abhijeet and Gwak, JunYoung and Frankel, Eric and Sadeghian, Amir and Savarese, Silvio},
    journal={IEEE transactions on pattern analysis and machine intelligence},
    year={2021},
    publisher={IEEE}
}

@inproceedings{yan2017online,
  title={Online learning for human classification in 3D LiDAR-based tracking},
  author={Yan, Zhi and Duckett, Tom and Bellotto, Nicola},
  booktitle={IEEE/RSJ Int. Conf. on Intell. Robots \& Systems (IROS)},
  pages={864--871},
  year={2017},
  organization={IEEE}
}

\end{document}